%% file: main.tex
\documentclass[lettersize, journal, twoside]{IEEEtran}

\IEEEoverridecommandlockouts                              % This command is only needed if 
\title{\LARGE \bf
\ourmethod{}: Cross-Embodiment Imitation Using a Grasping Tool
}

\author{Mingyo Seo$^{1}$$^{2}$, H. Andy Park$^{2}$, Shenli Yuan$^{2}$, Yuke Zhu$^{1}$$^{\dagger}$, and Luis Sentis$^{1}$$^{2}$$^{\dagger}$
\thanks{Manuscript received: August 22, 2024; Revised:
November 21, 2024; Accepted: January 7, 2025.
This paper was recommended for publication by Editor Jens Kober upon evaluation of the Associate Editor and Reviewers' comments.
This work was partially supported by the AI Institute
and the Office of Naval Research (N00014-22-1-2204).}
\thanks{
$^{1}$Mingyo Seo, Yuke Zhu, and Luis Sentis are with the University of Texas at Austin.
$^{2}$Mingyo Seo, H. Andy Park, Shenli Yuan, and Luis Sentis are with the AI Institute.
$^{\dagger}$Yuke Zhu and Luis Sentis contributed equally as advisers.} \thanks{Correspondance: {\tt\small mingyo@utexas.edu}}
\thanks{Digital Object Identifier (DOI): see top of this page.}
}
\let\labelindent\relax
\DeclareUnicodeCharacter{0301}{\'{e}}
\usepackage{times}
\usepackage{xcolor}

\usepackage{amsfonts}
\usepackage{amsmath}
\usepackage{amssymb}
\usepackage{graphicx}

\usepackage{booktabs}
\usepackage{hyperref}
\usepackage{dsfont}

\usepackage[font=small,labelfont=bf]{caption}
\usepackage{enumitem}
\usepackage[backend=biber,
            hyperref=true,
            url=false,
            isbn=false,
            doi=false,
            backref=false,
            style=ieee,
            citestyle=numeric-comp,
            sorting=none,
            block=none]{biblatex}
\usepackage[font=footnotesize]{caption}

\usepackage{algpseudocode}
\usepackage{algorithm}

\usepackage[switch]{lineno}

\setlist[itemize]{leftmargin=*}

\renewcommand{\bibfont}{\small}
\newcommand{\arxiv}[1]{{\textcolor{black}{#1}}}
\newcommand{\modify}[1]{{\textcolor{black}{#1}}}

\newcommand{\ourmethod}{{\textsc{LEGATO}}}
\newcommand{\ourtool}{{LEGATO Gripper}}
\newcommand{\task}[1]{\textbf{\textit{#1}}}
\begin{document}

\maketitle
% \thispagestyle{empty}
% \pagestyle{empty}

%%%%%%%%%%%%%%%%%%%%%%%%%%%%%%%%%%%%%%%%%%%%%%%%%%%%%%%%%%%%%%%%%%%%%%%%%%%%%%%%
\begin{abstract}

Cross-embodiment imitation learning enables policies trained on specific embodiments to transfer across different robots, unlocking the potential for large-scale imitation learning that is both cost-effective and highly reusable. 
This paper presents \ourmethod{}, a cross-embodiment imitation learning framework for visuomotor skill transfer across varied kinematic morphologies. 
We introduce a handheld gripper that unifies action and observation spaces, allowing tasks to be defined consistently across robots. 
We train visuomotor policies on task demonstrations using this gripper through imitation learning, applying transformation to a motion-invariant space for computing the training loss.
Gripper motions generated by the policies are retargeted into high-degree-of-freedom whole-body motions using inverse kinematics for deployment across diverse embodiments.
Our evaluations in simulation and real-robot experiments highlight the framework’s effectiveness in learning and transferring visuomotor skills across various robots.
More information can be found on the project page: {\tt\small\textbf{\url{https://ut-hcrl.github.io/LEGATO}}}.
\end{abstract}


%%%%%%%%%%%%%%%%%%%%%%%%%%%%%%%%%%%%%%%%%%%%%%%%%%%%%%%%%%%%%%%%%%%%%%%%%%%%%%%%
\input{introduction}
\input{related_work}
\input{method}
\input{experiments}
\input{conclusion}
% \newpage

{\footnotesize
\noindent
{\bf Acknowledgements}
This work was conducted during Mingyo Seo's internship at the AI Institute. 
We thank Rutav Shah and Minkyung Kim for providing feedback on this manuscript. 
We thank Osman Dogan Yirmibesoglu for designing the fin ray style compliant fingers and helping with hardware prototyping.
We thank Mitchell Pryor and Fabian Parra for their support with the real Spot demonstration.
}

% \clearpage
\renewcommand*{\bibfont}{\footnotesize}

% arxiv submission
\renewcommand{\baselinestretch}{0.96}
\printbibliography
\input{appendix}

% ral submission
% \renewcommand{\baselinestretch}{0.99}
% \printbibliography

\end{document}

%% file: introduction.tex
\section{Introduction}
\label{sec:intro}
\IEEEPARstart{R}{ecent} advancements in robot hardware—from wheeled manipulators to humanoid robots~\cite{kemp2022stretch, bd-spot, unitree-b1, agility-digit, fourier-gr1}—have greatly increased access to diverse robotic platforms. 
To fully leverage these advancements in supporting human activities, robots must autonomously perform a wide range of complex tasks.
Deep imitation learning has shown promise in training autonomous policies for sensorimotor skills, reducing the need for extensive human programming compared to traditional rule-based approaches. 
It has yielded impressive results in complex robotic systems~\cite{xie2020deep, seo2023trill} and across diverse dexterous manipulation tasks~\cite{zhao2023aloha, mandlekar2023mimicgen, sridhar2023memory}. 
However, such an approach typically requires demonstration data from a specific target robot, which limits scalability due to high hardware costs and the intensive workload involved in operating the robot during demonstrations.
Additionally, individualized training for each robot restricts cross-embodiment applications, as data cannot be transferred to different robot systems, even for similar tasks.

To enable scalable demonstration collection, pioneering works have introduced data collection tools that allow humans to directly manipulate during demonstrations~\cite{song2020grasping, young2021visual, pari2022vinn, seo2022prelude, chi2024umi}. 
These approaches enable training visuomotor policies that are deployable to specified target robots, reducing human workload and avoiding the costs and risks associated with using real robots for data collection.
However, these methods require either designing specialized data collection tools for specific robot gripper mechanisms or replacing robots' original grippers with customized hardware. 
This limitation restricts the applicability of these tools for robots with diverse gripper mechanisms.
For example, Universal Manipulation Interface~\cite{chi2024umi} is designed specifically for the Schunk WSG-50 gripper with its parallel-jaw mechanism but is incompatible with grippers employing other mechanisms, such as hinge types.
Additionally, variations in control latency and trajectory-tracking errors across robots, absent in human demonstrations, complicate policy transfer between embodiments. Chi et al.~\cite{chi2024umi} addressed this by compensating for control and observation latencies, while Song et al.~\cite{song2020grasping} used fine-tuning through trial and error on the target robot system. However, these strategies are difficult to generalize across diverse robots, as control latency varies between platforms due to differences in hardware and controllers.

\begin{figure}[!t]
	\centering
	\includegraphics[width=\linewidth]{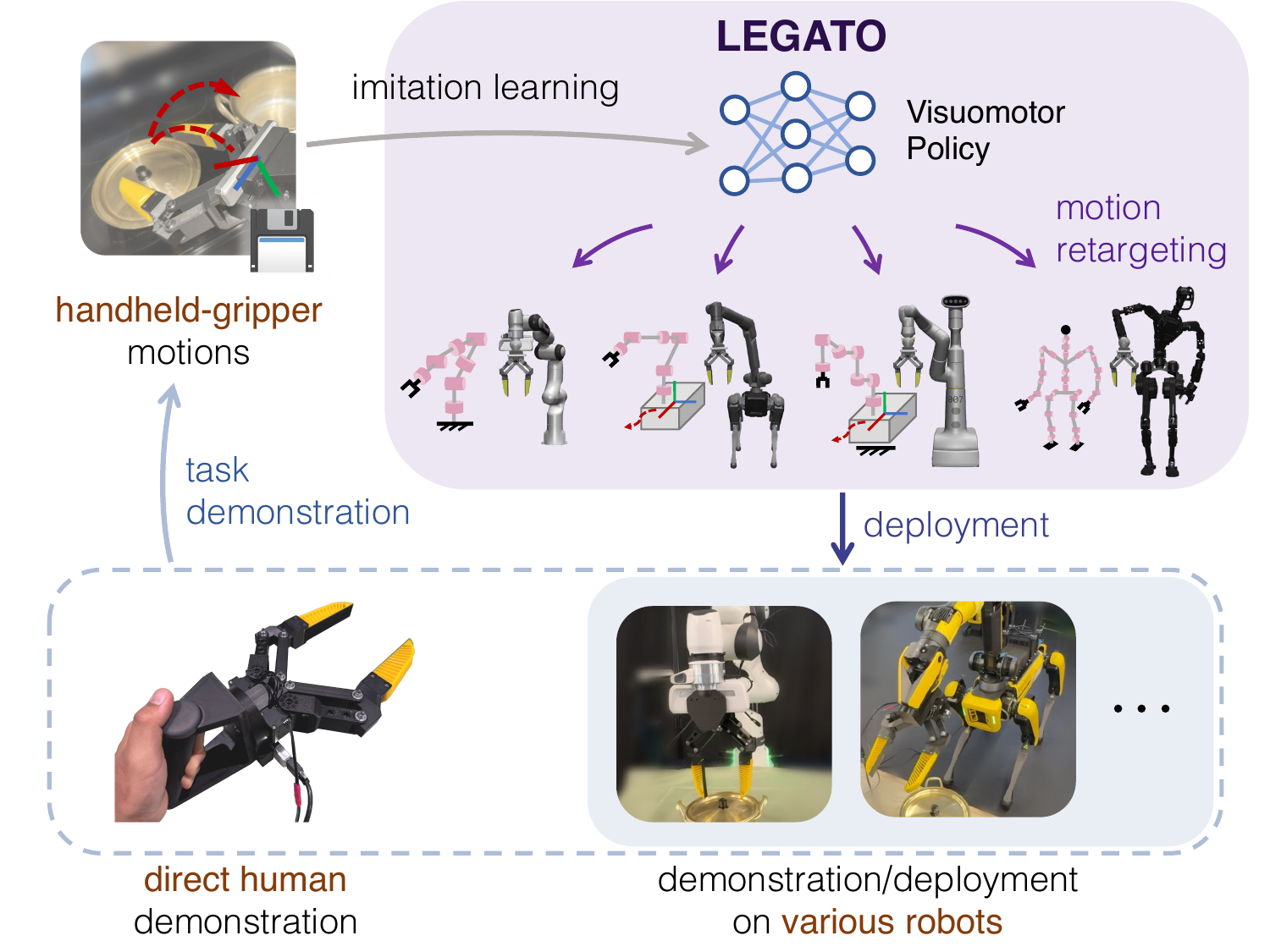}
	\caption
	{\textbf{Overview of \ourmethod{}.} 
\ourmethod{} addresses the challenge of transferring visuomotor skills across diverse robot embodiments. 
We present a cross-embodiment imitation learning framework using a versatile handheld grasping tool that ensures consistent physical interactions across different embodiments. 
Visuomotor policies trained on demonstrations by humans or teleoperated robots using the tool can be deployed across various robots equipped with the same gripper. 
Motion retargeting enables the execution of trajectories on different robots without requiring robot-specific training data.
}
	\label{fig:overview}
\end{figure}

Our key idea to address hardware differences across robots is to integrate a handheld grasping tool that can be shared across various robot embodiments for performing the same tasks (see Figure \ref{fig:overview}).
We name our method \ourmethod{} (\underline{L}earning with a Handh\underline{e}ld \underline{G}r\underline{a}sping \underline{To}ol). 
In our method, the \ourtool{}, a custom-designed handheld gripper, acts as a versatile and adaptive tool, representing tasks through its trajectories and grasping actions to ensure consistency across embodiments during both demonstration and deployment.
This handheld gripper—compatible with various robot grippers—enables a single visuomotor policy to be applied across diverse robot systems without requiring modifications to the original systems.
Sharing the gripper across robots and directly manipulating human demonstrators ensures consistent, actively actuated grasping actions.
In contrast, other data collection methods often create discrepancies in grasping actions because demonstrations rely on passive actuation by demonstrators, while robots require active actuation during deployment.

Our framework incorporates low-level motion retargeting through inverse kinematic (IK) optimization, tailored to each robot, along with a high-level transferable visuomotor policy (see Figure~\ref{fig:pipeline}).
Commands from the visuomotor policy for the handheld gripper are converted into whole-body motions using IK, enabling adaptation across robotic systems with only kinematic information, thereby avoiding extensive robot-specific training.
While the IK optimization adapts gripper motions to each robot, variations in hardware and kinematics introduce differences in control latency and errors.
To address this, we incorporate regularization on the gripper's trajectories in a motion-invariant space during training, preventing bias toward the demonstration embodiment and effectively learning motor skills. 
This approach ensures that gripper trajectories from the visuomotor policy can be consistently translated into whole-body robot motions, regardless of control latency and IK response differences.
Together, these components enable the learned visuomotor policy in \ourmethod{} to be effectively transferred across diverse robotic systems.

% Summary of key results
We validate our approach through simulation and real-robot experiments, demonstrating its cross-embodiment transferability. 
In simulation, visuomotor policies trained on human demonstrations are successfully deployed across various embodiments, including a tabletop manipulator, a wheeled robot, a quadruped, and a humanoid. 
We further demonstrate the reusability of demonstration data by transferring a policy trained on one robot to another. 
In real-world tabletop manipulator setups, our method achieves a 72\% success rate in complex manipulation tasks through policy transfer from direct human demonstrations.

%% file: related_work.tex
\section{Related Work}

\subsection{Skill Learning Across Embodiments}
Cross-embodiment policy transfer enables collecting demonstration data from easier or less costly embodiments and reusing it across different robot embodiments.
One approach involves learning from videos of direct human demonstrations, which has been extensively explored in earlier works~\cite{smith2019avid, zakka2022xirl, shaw2023videodex, wang2023mimicplay, xu2023xskill, okami2024}. 
However, robot deployment is limited by its reliance on third-person visual observations and the absence of real-world physical interaction data.

\begin{figure}[!t]
	\centering
	\includegraphics[width=0.95\linewidth]{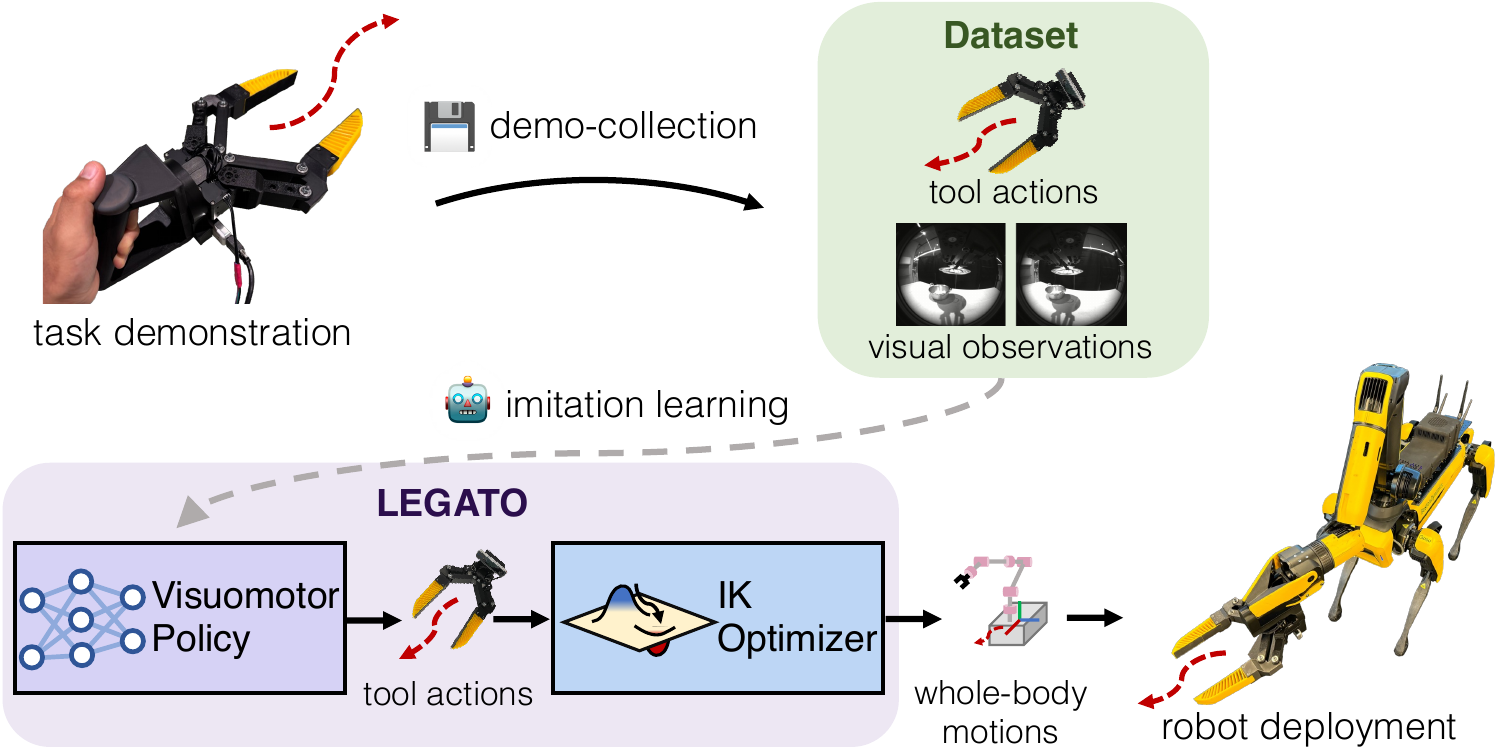}
	\caption
	{\textbf{\ourmethod{}'s cross-embodiment learning pipeline.} 
    During data collection, the \ourtool{} records its trajectories, grasping actions, and visual observations captured by its egocentric stereo camera.
    A visuomotor policy is then trained on these demonstrations through imitation learning.
    During deployment, the visuomotor policy's outputs are retargeted to the robots' whole-body motions through IK optimization.
    }
	\label{fig:pipeline}
\end{figure}

Another approach integrates tools specifically designed for demonstration collection.
One method involves leader-follower systems, which have been shown to be successful~\cite{fang2023airexo, kim2023training}. 
However, this method requires designing leader hardware with a kinematic structure identical to a target robot, enabling its joint states to be directly mapped onto the robot.
Alternatively, recent works use handheld tools to record on-hand visual observations and corresponding motions from demonstrations~\cite{song2020grasping, young2021visual, pari2022vinn, chi2024umi, ha2024umilegs}. 
While these approaches simplify tool design, they still require a system-specific tool or modifications to a robot's original hardware to align the end-effector’s physical interactions and the recorded visual observations.

Unlike prior methods, we aim to generalize cross-embodiment learning by incorporating an adaptable handheld gripper and flexible kinematics-based motion retargeting.
As a result, our learned policies are deployable across various types of robots without requiring robot-specific hardware for demonstrations.

\begin{figure*}[!t]
	\centering
	\includegraphics[width=0.9\linewidth]{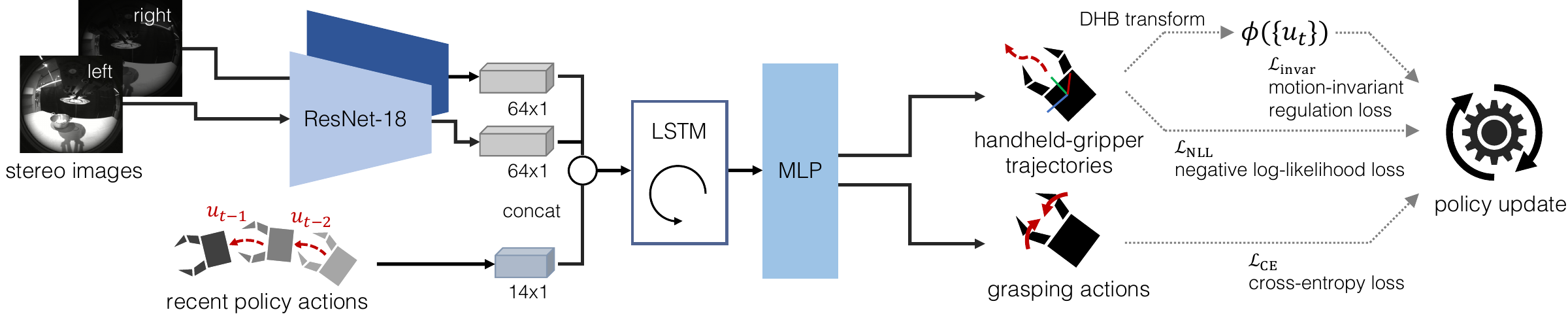}
	\caption
	{\textbf{High-level visuomotor policy architecture.}
    The trained policies generate desired handheld-gripper trajectories and grasping actions $u_t$ at 10 Hz from ego-centric stereo camera observations and previous policy actions. These action and observation spaces, defined in the handheld-gripper frame, remain consistent across various robot platforms. 
    To learn actions on handheld-gripper trajectories, we apply two action losses: the negative log-likelihood loss $\mathcal{L}_{\text{NLL}}$ for the distribution in SE(3) and the L2 loss $\mathcal{L}_{\text{invar}}$ in the DHB motion-invariant space.
    The grasping actions are trained using the cross-entropy loss $\mathcal{L}_{\text{CE}}$.
    }
	\label{fig:architecture}
\end{figure*}

\subsection{Whole-body Motion Retargeting}
Motion retargeting enables the generation of practical motions across different embodiments.
A substantial body of research exists on motion retargeting between similar morphologies, such as from humans to humanoid robots, employing either model-based methods~\cites{tachi2003telexistence, ayusawa2017motion, penco2018robust, darvish2019whole} or data-driven approaches~\cites{RoboImitationPeng20, he2024learning, fu2024humanplus}.
In contrast, translating human movements into equivalent motions for varying target morphologies presents a challenge due to the inherently ambiguous nature of the task.
Recent studies have investigated motion retargeting across embodiments with diverse morphologies~\cites{kim2022human, li2023ace, yan2023imitationnet}. 
While these studies have successfully demonstrated motion retargeting from humanoid to non-humanoid morphologies, they often rely on either embodiment-specific or task-specific models.

Our method shares similarities with previous works~\cite{penco2018robust, darvish2019whole} in utilizing kinematic optimization for motion retargeting.
Unlike these works, however, we use a lower-dimensional action space based on the motions of the handheld gripper. 
This approach enables flexible motion retargeting across diverse robot embodiments with varying morphologies.

\subsection{Trajectory Representations}
To effectively transfer motions across diverse embodiments, motion primitives offer a practical solution. 
These primitives facilitate the assembly of elemental motions, as extensively explored in the literature~\cite{saveriano2023dmp, paraschos2013promp, zhou2019vmp}. 
However, their adaptability is inherently limited by design, making them less effective for novel tasks or environments. 
Recent advancements, such as encoder-decoder frameworks that project trajectories into a latent space, have significantly improved the learning of motion primitives~\cite{noseworthy2020tc-vae, urain2020imitationflow, lee2023emmp}.
Encoding motions into learned latent spaces facilitates the transfer of human motions to simulated humanoid models~\cite{luo2023pulse} and adapts them to robot embodiments with varying kinematic structures~\cite{yan2023imitationnet}. 
Nevertheless, these approaches often encounter challenges in achieving broad generalization across different scenarios.

To address these challenges, our method employs motion representations as training regularization elements rather than directly using them for motion generation. 
Specifically, we utilize the Denavit-Hartenberg Bidirectional (DHB) invariant representation~\cite{lee2018dhb}, which offers invariance to rotational-translational shifts and scaling. 
The regularization in this motion-invariant space ensures robust alignment with demonstrated trajectories, enhancing policy generalizability across various embodiments.

%% file: method.tex
\section{Method}

Here, we introduce \ourmethod{}, a cross-embodiment imitation learning framework comprising a visuomotor policy at the high level and motion retargeting at the low level, as illustrated in Figure \ref{fig:pipeline}. 
The visuomotor policy is trained through imitation learning on task demonstrations, collected either from humans directly using the tool or from teleoperated robots holding the tool.
The consistency of the action space of handheld gripper motions and ego-centric visual observations across different robots enables deployment to various embodiments.
The low-level motion retargeting realizes these gripper trajectories as whole-body motions across different robot platforms.

\subsection{Problem Formulation}
We model the problem of cross-embodiment manipulation as a discrete-time Markov Decision Process $\mathcal{M}=(\mathcal{S}, \mathcal{A}, \mathcal{P}, R, \gamma, \rho_{0})$ where $\mathcal{S}$ is the state space, $\mathcal{A}$ is the action space, $\mathcal{P}(\cdot|s, a)$ is the transition probability, $R(s, a, s')$ is the reward function, $\gamma \in [0, 1)$ is the discount factor, and $\rho_{0}(\cdot)$ is the initial state distribution. Our objective is to learn a closed-loop visuomotor policy  $\pi(a_t|s_t)$ that maximizes the expected return $\mathbb{E}[\sum^\infty_{t=0}\gamma^t R(s_t, a_t, s_{t+1})]$ across different robot systems. In our problem, $\mathcal{S}$ is the space of visual observations captured by the handheld gripper's egocentric cameras, the robot's proprioceptive feedback, and previous actions, while $\mathcal{A}$ is the space of the robot's joint-space commands. These two spaces vary across different robot systems, reflecting the diversity in sensory capabilities and actuation mechanisms inherent to each robot platform. $R(s, a, s')$ is the reward function designed for the manipulation task, and $\pi$ is a closed-loop policy deployed on the robot. 

To handle the complexity of visuomotor skills and ensure that the policy $\pi$ is deployable across various robot systems, we decompose the policy into a two-level hierarchy. At the high level is a cross-embodiment visuomotor policy $\pi_{H}$, implemented as neural networks that compute target motion trajectories of the handheld gripper $u$. We train $\pi_{H}$ through imitation learning with demonstrations collected by manipulating the handheld gripper.
At the low level, we use a whole-body motion optimizer $\pi_{L}$, which determines target joint configurations to follow the trajectories $u$ established by $\pi_{H}$ through IK. This eliminates the need for additional training on target robot systems.
With this hierarchical structure, the entire policy can be represented as $\pi(a_t|s_t) = \pi_{L}(a_t|s_t, u_t)\pi_{H}(u_t|s_t)$.

\begin{figure*}
	\centering
	\includegraphics[width=\linewidth]{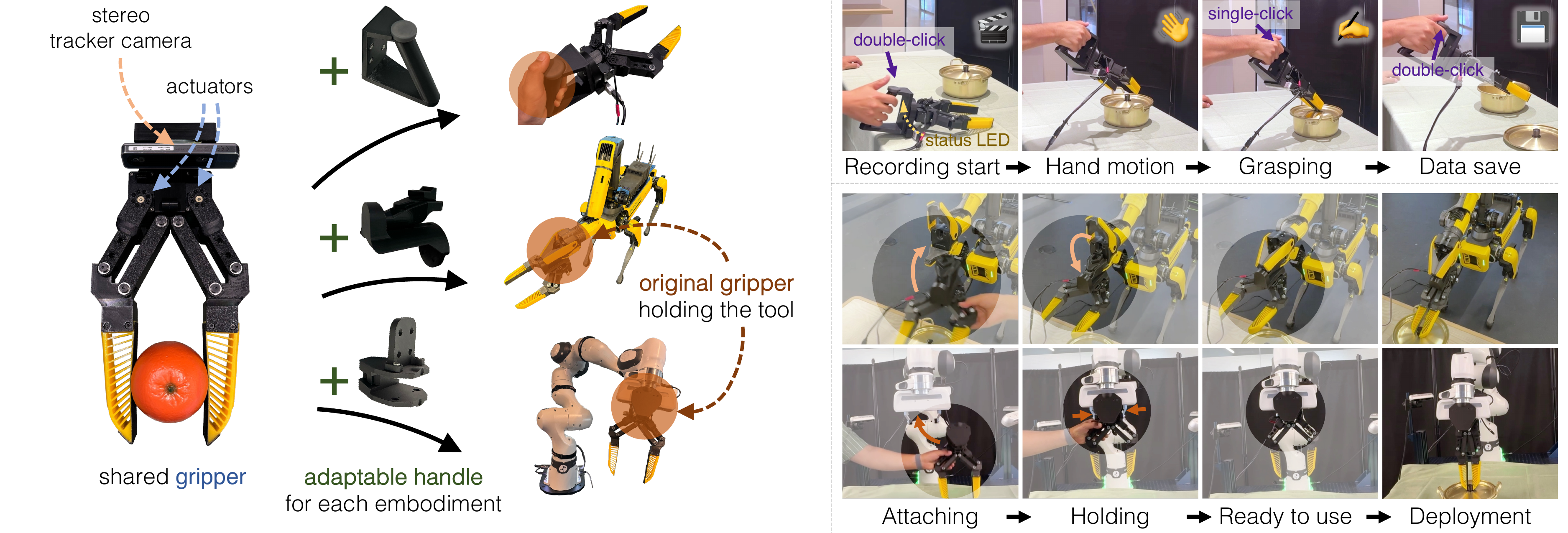}
	\caption
	{\textbf{\ourtool{} design.} 
    The \ourtool{} is designed for both human demonstration collection and robot deployment. 
    \textbf{(Left)} It features a shared actuated gripper with adaptable handles, ensuring reliable human handling and consistent grasping across robots while minimizing components. 
    \textbf{(Right top)} 
    A human demonstrator can directly perform tasks by carrying the \ourtool{} in hand. 
    The design includes a simple yet intuitive button interface with a status LED, allowing data recording to start and end with a double-click and grasping actions to trigger with a single click.
    \textbf{(Right bottom)} The \ourtool{} is easily installed on various robots, securely held by their original grippers, and is ready for immediate use.
    }
	\label{fig:gripper}
\end{figure*}

\subsection{Actions Based on the Handheld Gripper}
\label{sec:action}
We incorporate the \ourtool{}, which can be shared across different robot systems. 
Inspired by Noguchi et al.~\cite{noguchi2021tool}, each robot uses its own gripper to hold the handheld gripper, integrating it as part of the embodiment. 
The handheld gripper maintains a consistent shape and viewpoint, unifying visual observation and action spaces across embodiments. 
This reduces the complexity of the cross-embodiment problem, streamlining the mapping of handheld-gripper motions and coordinating whole-body motions to execute them.

We define the action space as the differential pose in SE(3) between consecutive time steps. 
This action space is suitable for generalizable whole-body manipulation on floating-base robots, as it eliminates reliance on a fixed reference frame. 
Actions are thus represented in the current handheld-gripper frame. 
Differential pose actions, sampled from a Gaussian Mixture Model~\cite{wang2020critic}, capture the multimodal nature of human demonstrations and essential motion information within the motion-invariant space, as described in Section \ref{sec:policy}.

\subsection{Whole-body Motion Retargeting}
In this section, we detail our approach for mapping the trajectory actions of the handheld gripper $u_t$ into whole-body robot motions $a_t$. We employ an optimization-based IK method to handle the constrained IK problem. 
This enables our motion optimizer $\pi_{L}$ to map commands effectively to robot motions by leveraging degree-of-freedom (DOF) redundancy while respecting actuation bounds and other constraints.
Our approach addresses kinematic differences, constraints, and the diversity in DOFs across robot embodiments by using kinematic redundancy to satisfy joint and other constraints during deployment, without requiring additional robot-specific demonstrations.

The motion retargeting formulation is expressed as the following quadratic programming (QP) problem:
\begin{equation}
\label{eq:optimization}
\begin{aligned}
& \min_{a_t}
\sum {\frac{1}{2}} a_t^{\top} {H} {a_t} \\
& \text{subject to} \\
& \quad J_i(q_{t})a_{t} = \dot{x}_i^{\text{des}}, \\
& \quad \underline{L}_{i} \leq C_i a_t \leq \overline{L}_{i},
\end{aligned}
\end{equation}
where 
$\underline{L}_{i}$, $\overline{L}_{i}$ and $C_i$ define the velocity-level constraints for joint positions $q$, velocities $\dot{q}$, and accelerations $\ddot{q}$, along with other Cartesian limits (e.g., virtual walls and collision distance bounds). $H$ is a positive-definite weighting matrix.  
$\dot{x}_i^{\text{des}}$ represents the desired velocities at the prioritized tasks, formulated in the following order:
\begin{equation}
\begin{aligned}
\label{eq:ik_task}
&J_1 = J_{\text{grip}}, \quad \dot{x}_1^{\text{des}} = K_{\text{grip}}(x_{\text{des}} - x(q)), \\
&J_2 = I_{n_j}, \quad \dot{x}_2^{\text{des}} = K_{\text{bias}}(q_{\text{bias}} - q),
\end{aligned}
\end{equation}
where $J_{\text{grip}}$ and $I_{n_j}$ represent the Jacobians for gripper pose control and maintaining the configuration bias $q_{\text{bias}}$ of $n_j$ DOFs, respectively, with the control gains $K_{\text{grip}}$ and $K_{\text{bias}}$.
% Additionally, $x(q_t)$ and $J_{\text{eef}}(q_t)$ represent the pose and the task-space Jacobian matrix of the end-effector at joint state $q_t$, while $\Delta t$ denotes the period of motion retargeting.
${x}_{\text{des}}$ represents the target pose of the handheld gripper, as determined by the visuomotor policy outputs $u_t$.

To solve the QP problem with the hierarchical priorities and the inequality constraints, we utilize the extended Saturation in the Null Space (eSNS) algorithm~\cite{fiore2023general}. 
The constrained optimization problem of Equation \ref{eq:optimization} is formulated as follows:
\begin{equation}
\begin{aligned}
& \text{maximize} \quad \sum_{i} c_i\\
& \text{subject to} \\
&\quad J_i(q_{t})a_{t} = c_i \dot{x}_i^{\text{des}}, \\
& \quad \underline{L}_{i} \leq C_i a_t \leq \overline{L}_{i}.
\end{aligned}
\end{equation}
Here, $c_i$ represents scaling factors within the range [0,1] that allow for scaling speed of the prioritized tasks defined in Equation \ref{eq:ik_task}, while accurately tracking their trajectories. 
This optimization problem, maximizing the sum of scaling factors, yields the closest mapping motions while meeting all constraints. 
Our framework ensures that robot movements not only follow the prescribed hierarchy but also adhere to the constraints, guaranteeing flexible task execution and robust motion generalization.

\begin{figure*}[!t]
	\centering
	\includegraphics[width=\linewidth]{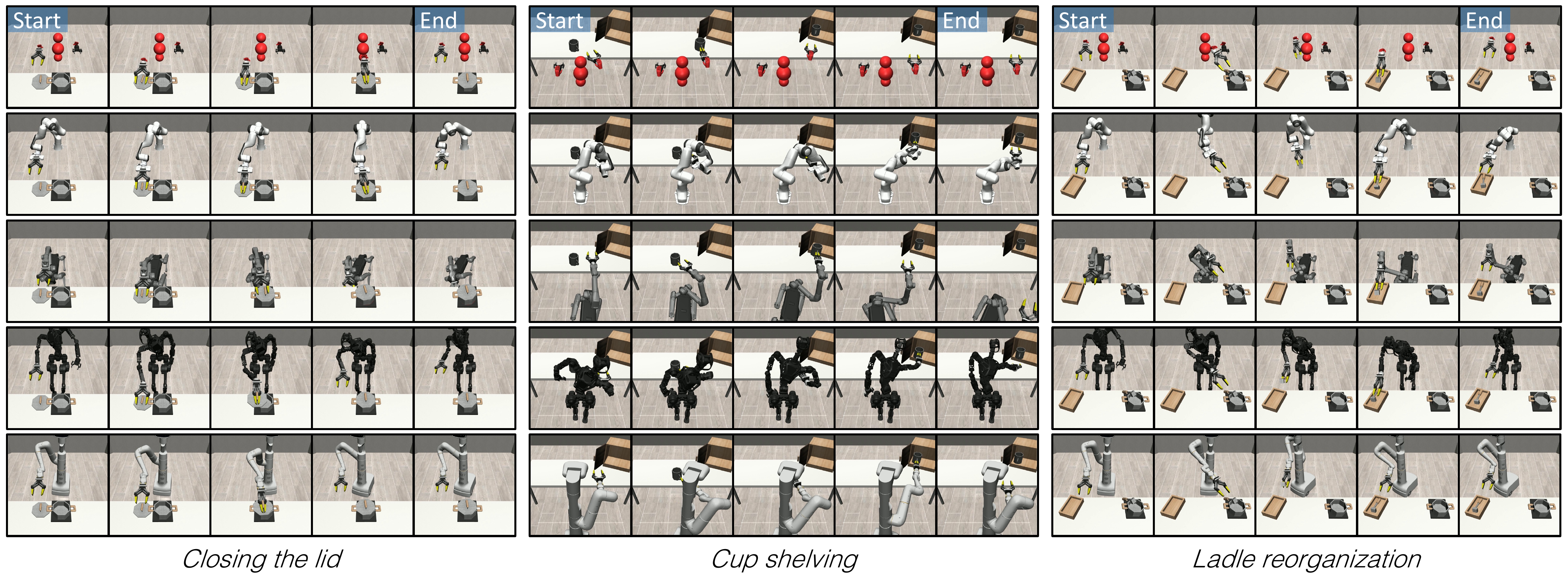}
	\caption{
    \textbf{Timelapse of deploying \ourmethod{} in simulation.}
        We trained visuomotor policies using demonstrations from the \textit{Abstract} embodiment and deployed them on robots with diverse morphologies, from the top: \textit{Abstract}, \textit{Panda}, \textit{Spot}, \textit{GR-1}, and \textit{Google Robot}. The timelapse of deploying these policies reveals consistent time steps.
        The tracking performance of the IK motion retargeting varies with morphology, leading to domain gaps across embodiments. 
        Despite these challenges, \ourmethod{} achieves successful deployment on various robots.
        }
	\label{fig:timelapse_sim}
\end{figure*}

\subsection{Training of Visuomotor Policies}
\label{sec:policy}
Task demonstrations can be performed on any embodiment capable of executing the tasks within its joint space.
The collected demonstration dataset $\mathcal{D}$ consists of state-action pairs ${\mathcal{D}=\{(s_i, u_i)\}_{i=1}^N}$.
Here, $N$ represents the total number of data points.
The observations $s_i$ comprise stereo images from the handheld gripper's onboard camera and a history buffer of previous actions ${u_k}_{k=t-T:t-1}$.
The visual observations are provided as stereo grayscale images with a resolution of $128 \times 128$ pixels.
The demonstration commands $u_i$ include the subsequent setpoints for the handheld gripper (6D) and the grasping actions (1D).

We train our policy $\pi_H$ using a deep imitation learning algorithm, specifically using a behavioral cloning policy with LSTM networks~\cite{hochreiter1997long, mandlekar2021matters}, as shown in Figure~\ref{fig:architecture}. 
The visuomotor policy employs two separate ResNet18-based image encoders~\cite{he2016deep}, trained end-to-end. 
The encoded features are flattened and processed by two-layer LSTM networks, each with 400 hidden units.
The policy outputs are generated by a three-layer Multi-Layer Perceptron (MLP), with each layer containing 2048 hidden units. 
For hand trajectories, the policy outputs Gaussian Mixture Model (GMM) parameters to determine the next target pose based on spatial and rotational differences from the previous frame, using a 5-mode GMM. 
Grasping actions are provided as binary classifications for opening and closing the gripper.

For imitation learning, we employ behavioral cloning with the following training loss:

\begin{equation}
\label{eq:loss}
\begin{aligned}
\mathcal{L} = \mathcal{L}_{\text{NLL}} + \mathcal{L}_{\text{invar}} + \mathcal{L}_{\text{CE}}.
\end{aligned}
\end{equation}
Here, $\mathcal{L}_{\text{NLL}}$ and $\mathcal{L}_{\text{invar}}$, associated with the trajectories of the handheld gripper, represent the negative log-likelihood loss and the regularization term in the motion-invariant space, respectively. 
Additionally, $\mathcal{L}_{\text{CE}}$ denotes the cross-entropy loss for binary grasping actions. 
The term $\mathcal{L}_{\text{invar}}$ is calculated as the L2 loss in the motion-invariant space, specifically using the DHB invariant transformation~\cite{lee2018dhb}, $\phi(\cdot)$\arxiv{\ (see Appendix for details)}. This transformation is applied to the differential poses of the gripper in SE(3) from the policy outputs $u_t$ and the demonstrations $\hat{u}_t$:
\begin{equation}
\begin{aligned}
\label{eq:invar}
\mathcal{L}_{\text{invar}} = \sum \lVert
\phi(u_t, \{\hat{u}_{k}\}_{k=t-T}^{t-1}) - \phi(\{\hat{u}_{k}\}_{k=t-T}^{t})
\rVert ^2.
\end{aligned}
\end{equation}
Incorporating $\mathcal{L}_{\text{invar}}$ into the training loss influences the distribution of handheld-gripper motions, preventing the policy from being biased by embodiment-specific properties like tracking errors and control latency.
Unlike SE(3), these invariants represent motion by breaking it down into magnitude and directional changes, unaffected by embodiment-specific factors such as viewpoints, reference frames, pose offsets, or scales.
Leveraging motion invariance ensures the policy captures essential motion information from demonstrations without bias toward specific embodiments.
This is critical for cross-embodiment learning, as it makes the policy robust to domain mismatches across different robot embodiments.

%% file: experiments.tex
\section{Experiments}
\label{sec:exps}
In this section, we demonstrate the feasibility and effectiveness of \ourmethod{} for cross-embodiment transfer of visuomotor policies, both in MuJoCo simulation~\cite{todorov2012mujoco} and real-world settings. Leveraging the scalability and ease of simulation environments, we employ them to investigate the following research questions:
1) How does the regularization in the motion-invariant space impact the training of cross-embodiment policies?
2) How do differences in morphology and controllability affect the task capacity of robot embodiments?

\begin{figure*}[!t]
	\centering
	\includegraphics[width=\linewidth]{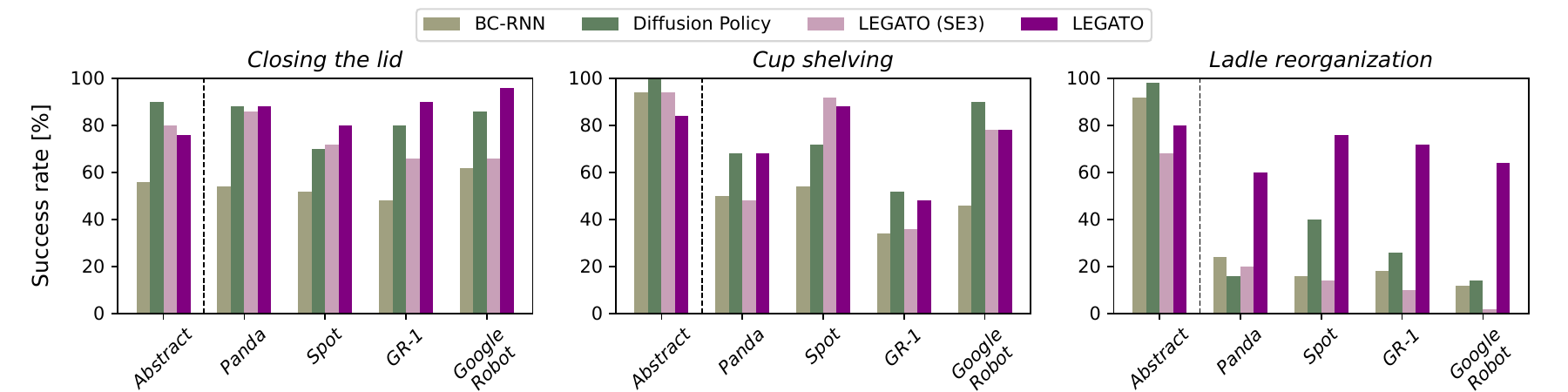}
	\caption{
    \textbf{Quantitative results in simulation.}
    We report success rates on 50 trials of our \ourmethod{} policies compared to baselines. 
    On average, \ourmethod{} outperforms all other methods in cross-embodiment deployment by 28.9\%, 10.5\%, and 21.1\%, compared to BC-RNN, Diffusion Policy, and \ourmethod{} (SE3), respectively.
    Notably, unlike the baselines that achieve high success rates only on specific robot embodiments, typically the \textit{Abstract} embodiment used for training, \ourmethod{} demonstrates consistent success rates across different embodiments.
    }
	\label{fig:benchmark}
\end{figure*}

\subsection{\ourtool{}}
The \ourtool{} design described in Section \ref{sec:action} is implemented on real hardware.
It is designed for intuitive use in both direct human demonstrations and robot operations, as illustrated in Figure \ref{fig:gripper}. 
During human demonstrations, a user carries the \ourtool{} and controls the grasping actions. 
Robots, on the other hand, can hold the tool with their original grippers without needing any hardware modifications.

To enable easy attachment and replacement, the design is modularized, particularly the handle parts that robots grip, as presented in Figure \ref{fig:gripper}. 
Thus, only the handle parts need replacement for different robots, allowing all other core components to be shared across robots.
\modify{These handle parts can be easily designed using CAD models of target robots provided by manufacturers, ensuring adaptability to diverse robot systems.}
The \ourtool{} features two pairs of parallel four-bar linkage mechanisms, each actuated by one DOF, enabling a wider and more flexible range of opening distances.
The tool employs fingertips of compliant fin ray mechanisms for high compliance and adaptability during contact.
Made from 3D-printed parts, with TPU 95A for the fingertips and PLA for other components, the \ourtool{} is significantly lighter than typical commercial grippers.
These key features facilitate the concept of integrating the tool as part of the embodiment, even under robots' limited payload capacities, while supporting generalizable grasping actions.

The \ourtool{} is equipped with a Realsense T265~\cite{realsense2022} camera (or an alternative stereo tracking camera such as the SeerSense XR50~\cite{seersense2024}), with fisheye stereo cameras and an IMU for streaming visual observations and estimating the handheld gripper's motions via visual odometry.
During demonstrations, both stereo images and visual odometry data are recorded to form observation-action pairs.
In contrast, during robot deployment, only stereo images are streamed.

In simulation, the mechanism and geometry of the real hardware are modeled. Visual observations are emulated by adapting the properties of the tracker camera used in the hardware design. The \ourtool{} is attached to each robot's original gripper with an offset.

\subsection{Experimental Setup}
We designed the following three realistic manipulation tasks to study the cross-embodiment deployment of our method for both simulation and real robot systems.
\begin{itemize}
    \item \task{Closing the lid}\textbf{:} A robot grasps the lid and places it on the pot within reach. This task requires precise manipulation to accurately close the lid.
    \item \task{Cup shelving}\textbf{:} 
    A robot places the cup into the shelf. In addition to requiring a large workspace, this task involves complex collision-free motion to position the cup against the shelf's non-convex shapes.
    \item \task{Ladle reorganization}\textbf{:} A robot picks up the ladle from the pot and places it into the utensil organizer. This task involves complex manipulation and handling objects potentially out of view due to limited visibility and occlusion.
\end{itemize}
A task is considered successful if a robot accomplishes the designated goals within a specific time limit.
Across all baselines and tasks, the initial states of robots are consistent, and the initial poses of objects, other than the table, are uniformly randomized.

\subsection{Quantitative Evaluation in Simulation}
\label{sec:sim-eval}
In these experiments, we demonstrate the effectiveness of our domain-transfer method across different embodiments.
To systematically evaluate our method, we use embodiments representing various kinematic morphologies, as shown in Figure \ref{fig:overview}: Franka Emika \textit{Panda}~\cite{franka-panda}, Boston Dynamics \textit{Spot}~\cite{bd-spot}, Fourier \textit{GR-1}~\cite{fourier-gr1}, and \textit{Google Robot}~\cite{herzog2023deep}.
\begin{itemize}
    \item \task{Abstract}\textbf{:} an idealized embodiment designed for simulation that can manipulate hands along continuous trajectories without speed or workspace limitations. Human motion commands are directly mapped to the simulation, replicating direct human demonstration in real-world settings.
    \item \task{Panda}\textbf{:} a 7-DOF tabletop manipulator, used to demonstrate the impact of redundant DOFs on motion retargeting compared to the robots listed below.
    \item \task{Spot}\textbf{:} a quadrupedal robot with 6 DOFs in its arm and 6 DOFs in its body pose. To demonstrate achieving an extensive workspace through whole-body motion alone, the robot’s locomotion is not considered; instead, the leg joints track the robot's body within a limited SE(3) range.
    \item \task{GR-1}\textbf{:} a humanoid robot with 7 DOFs per arm and 3 DOFs each for the head and torso. Similar to \textit{Spot}, the robot’s locomotion is not considered, and the leg joints are fixed. This demonstrates the application of our method to highly redundant DOF systems.
    \item \task{Google Robot}\textbf{:} a wheeled mobile robot with 7 DOFs in its arm and 3 DOFs in planar base motion, used to show how handheld-gripper trajectories map to mobile manipulation.
    \end{itemize}
We focus on demonstrations from the \textit{Abstract} embodiment, where commands from human demonstrators are directly mapped, enabling task performance without being affected by tracking errors that occur in robots. 
The trained policies are then transferred directly to the robots without any adaptation, using the pre-defined IK optimizer, as shown in Figure~\ref{fig:timelapse_sim}.

Figure~\ref{fig:benchmark} reports quantitative evaluations of our simulated tasks, comparing our model with the following baselines.
\begin{itemize}
    \item \textbf{BC-RNN}\textbf{:} a baseline that uses recurrent neural networks to learn manipulation skills from teleoperated demonstrations, as introduced by Mandlekar et al.~\cite{mandlekar2021matters}. It employs flat GMM outputs and is trained with the negative log-likelihood loss.
    \item \textbf{Diffusion Policy}\textbf{:} a baseline that employs a receding-horizon visual-motor policy based on a diffusion model, as introduced by Chi et al.~\cite{chi2024diffusionpolicy}. In particular, we adapt the Velocity Diffusion Policy. Due to the requirement for precise tracking and state estimation of the end-effector pose with respect to the global frame, the Position Diffusion Policy is not applicable in our settings.
    \item \textbf{\ourmethod{} (SE3)}\textbf{:} a variant of our final model that excludes the motion-invariant regularization loss $\mathcal{L}_{\text{invar}}$ from the training loss, with all other components unchanged. This variant is designed to evaluate the impact of the regularization in the motion-invariant space on cross-embodiment deployment.
\end{itemize}
All baselines are trained on the same dataset of 150 task demonstrations using the \textit{Abstract} embodiment with identical policy inputs, as shown in Figure~\ref{fig:architecture}.
They output handheld gripper trajectories realized by the same IK optimizer, using consistent parameters within a single robot.

As shown in Figure~\ref{fig:benchmark}, \ourmethod{} outperforms the three baseline methods across various robot embodiments. 
The success rates for the \textit{Abstract} embodiment indicate that all baselines are well-trained within that domain. 
However, the higher success rates of \ourmethod{} (SE3) and Diffusion Policy, which uses the state-of-the-art diffusion model and performs best in the \textit{Abstract} domain, are followed by a notable decline when applied to other robots.
This suggests that exclusive training in SE(3) may introduce domain bias toward the training embodiment.
In contrast, the performance improvement of \ourmethod{} across diverse embodiments suggests that its regularization in the motion-invariant space enhances policy robustness across different robot domains, even with a simple model architecture. 
% The higher success rates of \ourmethod{} (SE3) in the \textit{Abstract} embodiment, followed by a significant drop in other robots, indicate that training solely in SE(3) can introduce domain bias toward the domain of the training embodiment.
% The comparison between \ourmethod{} (SE3) and BC-RNN highlights that tailoring the action space for grasping and handheld gripper trajectories improves performance. 
In addition, \ourmethod{} and the baselines show reduced performance overall in the \textit{Cup shelving} task with the \textit{Panda} and \textit{GR-1} robots, as well as in the \textit{Ladle reorganization} task with the \textit{Panda} robot.
This underscores the limitations of fixed-base robots for tasks requiring extensive motion, highlighting the need for whole-body manipulation.
\arxiv{Supplementary evaluations for further discussion are provided in Appendix.}

\begin{figure}[!t]
	\centering
	\includegraphics[width=\linewidth]{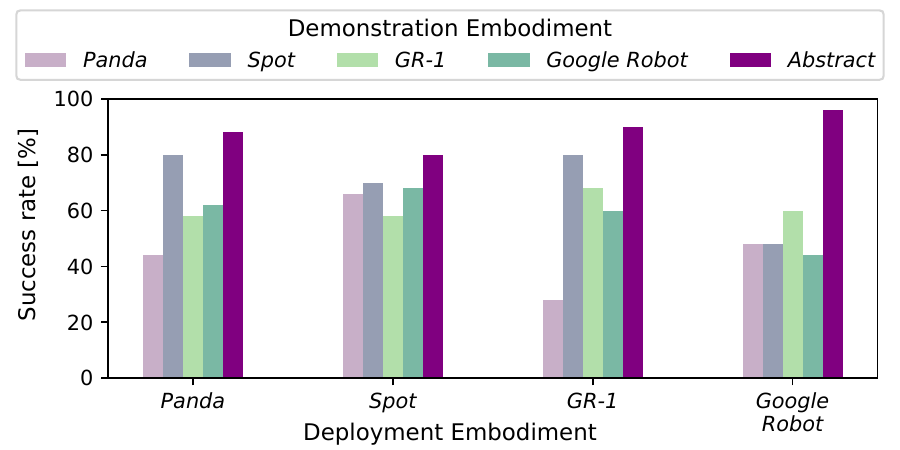}
	\caption{
    \textbf{Simulation evaluation across different demonstration embodiments.}
    We show success rate changes on 50 trials from policies trained on demonstrations of different embodiments.
    }
	\label{fig:embodiment}
\end{figure}

\subsection{Varying Demonstration Embodiments}
We investigate how different embodiments used in demonstrations affect policy performance in simulation. 
We collected 150 trajectories by teleoperating each robot in Section~\ref{sec:sim-eval} for the \textit{Closing the lid} task and trained policies on these demonstrations. 
Figure~\ref{fig:embodiment} shows the success rates for deploying these policies across the robots.
The policies trained on the \textit{Abstract} embodiment performed best, indicating that joint controller latency and the IK motion retargeting affect demonstration quality. 
Among the policies trained on robot teleoperation, demonstrations on the \textit{Spot} robot yield the highest success rates, whereas those on the \textit{Panda} robot yield the lowest. 
This suggests that whole-body motion capability and redundancy influence task demonstration quality, as the \textit{Spot} robot has full 6 DOFs in its body, while the \textit{Panda} robot has less redundancy.
Notably, policies trained on the \textit{Panda} and \textit{Google Robot} exhibit lower success rates when deployed on the robot used for demonstrations compared to other robots. 
This highlights that the trajectory tracking capabilities constrained by the deployment embodiments affect task complexity.

\begin{figure}[!t]
	\centering
	\includegraphics[width=\linewidth]{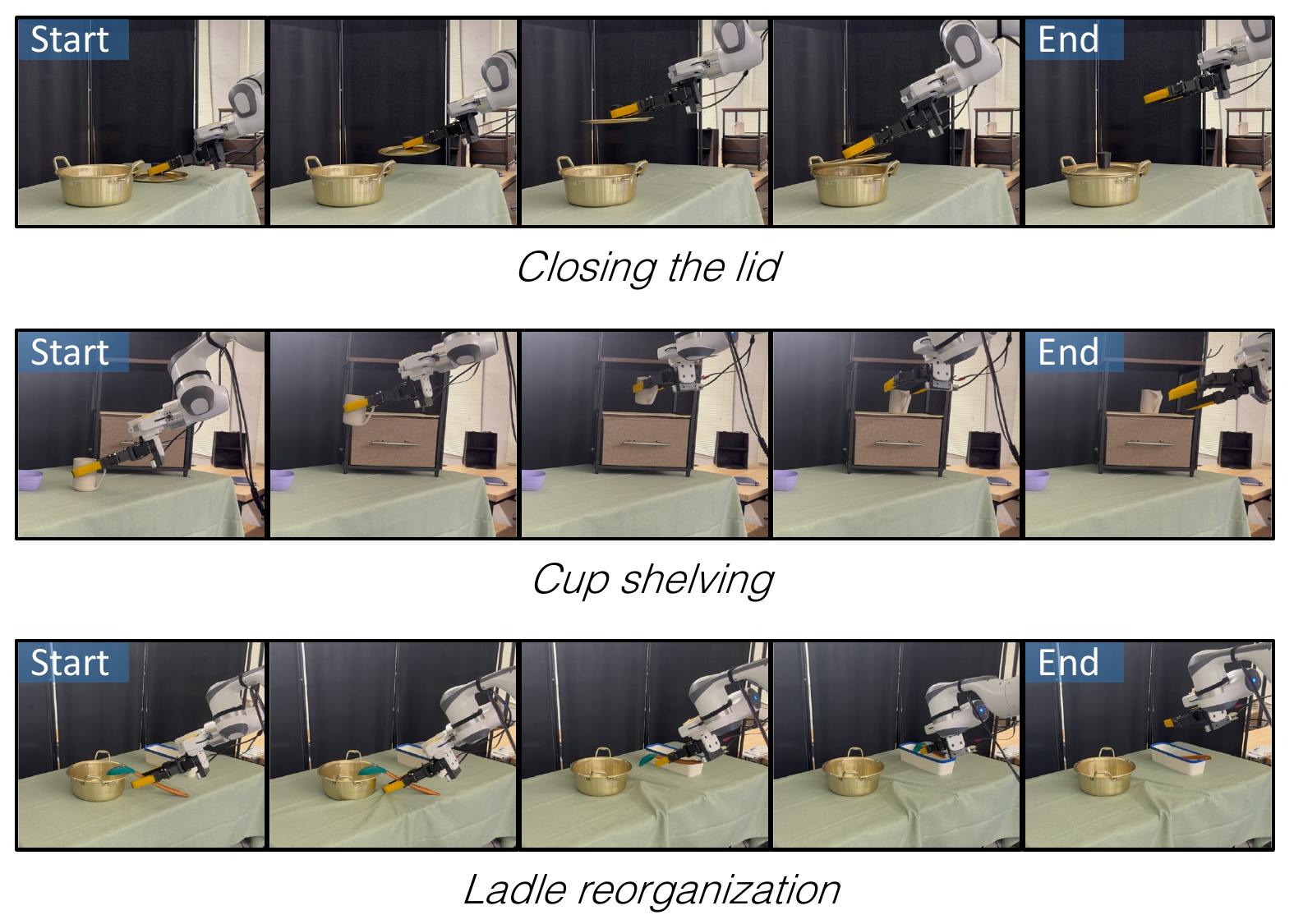}
	\caption{
    \textbf{Timelapse of deploying \ourmethod{} in the real-robot system.}
    We trained visuomotor policies on direct human demonstrations and successfully deployed them on the \textit{Panda} robot system. 
    The timelapse of deploying these policies demonstrates consistent time intervals.
    }
	\label{fig:timelapse_real}
\end{figure}

\subsection{Real-Robot Experiments}
\label{sec:real-eval}
We aim to validate the robustness of our method against variations caused by controller limitations and sensor inaccuracies in real-world settings. 
We collected direct human demonstrations through the \ourtool{} to train policies and deployed them, particularly on the \textit{Panda} robot, the most challenging due to its limited workspace at the IK motion retargeting, as shown in Section \ref{sec:sim-eval}.
During the evaluation of each task, the same visuomotor policy was attempted for 20 trials on the \textit{Panda} robot. 
Our method succeeded in 16 trials of the \textit{Closing the lid} task, 13 trials of the \textit{Cup shelving} task, and 14 trials of the \textit{Ladle reorganization} task, respectively (see Figure~\ref{fig:timelapse_real}).
\arxiv{More videos are provided on our project website.}

%% file: conclusion.tex
\section{Conclusion}
\label{sec:conclusion}

We present \ourmethod{}, a cross-embodiment learning framework for transferring visuomotor skills across diverse robot morphologies. 
By using a handheld gripper for consistent observations and actions, our framework enables visuomotor policies to transfer across embodiments without hardware modifications. 
Handheld-gripper trajectories from the visuomotor policies are mapped to whole-body robot motions through IK optimization. 
Although trained on specific embodiments, regularization in a motion-invariant space allows these policies to adapt easily to different robots, managing variations in control latency and tracking errors. 
Our current focus is on whole-body manipulation, employing redundant DOFs for flexibility and an extended workspace through coordinated body movements, though our method is limited to non-walking scenarios.
Future work will incorporate loco-manipulation, integrating manipulation with walking to enable legged robots to perform diverse tasks with larger workspaces. 
\modify{Additionally, while our method currently relies on designing handles specific to target robots, this can be addressed by identifying grasping location to enable the use of universal handles across robots.}
We also believe our approach is adaptable to a wide range of tools and applications beyond the \ourtool{} design.

%% file: appendix.tex
\section* {Appendix}
\label{sec:appendix}

\renewcommand\thesubsection{\Alph{subsection}}

\subsection{Motion-Invariant Regulation Loss}
\label{sec:invar}

The motion-invariant transform $\phi(\cdot)$, used to compute $\mathcal{L}_{\text{invar}}$ in Equation~\ref{eq:invar}, follows the DHB motion-invariant framework~\cite{lee2018dhb}. Given trajectories $\{{u}_{k}\}_{k=t-T}^{t}$ with $T \geq 2$, we compute the relative positions ${p}_{k}$ and orientations ${r}_{k}$ of the gripper with respect to the initial frame at ${t-T}$, where ${p}_{t-T}$ and ${r}_{t-T}$ are at the origin.

The differences $\Delta \mathbf{p}_k = \mathbf{p}_{k+1} - \mathbf{p}_k$ and $\Delta \mathbf{r}_k = \mathbf{r}_{k+1} - \mathbf{r}_k$ represent the linear and angular trajectory changes between $k+1$ and $k$. The initial linear frames are defined as:
\begin{equation*}
\begin{aligned}
    \hat{\mathbf{x}}_{p,k} &= \frac{\Delta \mathbf{p}_k}{\|\Delta \mathbf{p}_k\|},\\
    \hat{\mathbf{y}}_{p,k} &= \frac{\hat{\mathbf{x}}_{p,k} \times \hat{\mathbf{x}}_{p,k+1}}{\|\hat{\mathbf{x}}_{p,k} \times \hat{\mathbf{x}}_{p,k+1}\|},\\
    \hat{\mathbf{z}}_{p,k} &= \hat{\mathbf{x}}_{p,k} \times \hat{\mathbf{y}}_{p,k}.
\end{aligned}
\end{equation*}
Similarly, the initial angular frames are:
\begin{equation*}
\begin{aligned}
    \hat{\mathbf{x}}_{r,k} &= \frac{\Delta \mathbf{r}_k}{\|\Delta \mathbf{r}_k\|},\\
    \hat{\mathbf{y}}_{r,k} &= \frac{\hat{\mathbf{x}}_{r,k} \times \hat{\mathbf{x}}_{r,k+1}}{\|\hat{\mathbf{x}}_{r,k} \times \hat{\mathbf{x}}_{r,k+1}\|},\\
    \hat{\mathbf{z}}_{r,k} &= \hat{\mathbf{x}}_{r,k} \times \hat{\mathbf{y}}_{r,k}.
\end{aligned}
\end{equation*}
The directions of the axes in both frames are chosen to prevent discontinuities across time steps.

In the DHB transformation, the motion of a rigid body is separated into position and orientation components using two frames. Two invariants are the norms of the relative positions and orientations between these frames:
\begin{equation*}
\begin{aligned}
    m_{p,k} &= \|\Delta \mathbf{p}_k\|,\\
    m_{r,k} &= \|\Delta \mathbf{r}_k\|.
\end{aligned}
\end{equation*}
These invariants, $m_p$ and $m_r$, describe the translation of the linear and angular frames. Four additional values describe their rotation:
\begin{equation*}
\begin{aligned}
    \theta^1_{p,k} &= \arctan \left( \frac{\hat{\mathbf{x}}_{p,k} \times \hat{\mathbf{x}}_{p,k+1}}{\hat{\mathbf{x}}_{p,k} \cdot \hat{\mathbf{x}}_{p,k+1}} \cdot \hat{\mathbf{y}}_{p,k} \right),\\
    \theta^2_{p,k} &= \arctan \left( \frac{\hat{\mathbf{y}}_{p,k} \times \hat{\mathbf{y}}_{p,k+1}}{\hat{\mathbf{y}}_{p,k} \cdot \hat{\mathbf{y}}_{p,k+1}} \cdot \hat{\mathbf{x}}_{p,k+1} \right),\\
    \theta^1_{r,k} &= \arctan \left( \frac{\hat{\mathbf{x}}_{r,k} \times \hat{\mathbf{x}}_{r,k+1}}{\hat{\mathbf{x}}_{r,k} \cdot \hat{\mathbf{x}}_{r,k+1}} \cdot \hat{\mathbf{y}}_{r,k} \right),\\
    \theta^2_{r,k} &= \arctan \left( \frac{\hat{\mathbf{y}}_{r,k} \times \hat{\mathbf{y}}_{r,k+1}}{\hat{\mathbf{y}}_{r,k} \cdot \hat{\mathbf{y}}_{r,k+1}} \cdot \hat{\mathbf{x}}_{r,k+1} \right).
\end{aligned}
\end{equation*}
This process produces the linear and angular invariant values $(m_{p,k}, \theta_{p,k}^1, \theta_{p,k}^2)$ and $(m_{r,k}, \theta_{r,k}^1, \theta_{r,k}^2)$, as established in the original work.

To ensure continuity, the computed frame rotations are transformed using $\sin(\cdot)$ and $\sin(2\cdot)$. The final transformation applied in our regularization loss thus yields 10 variables of length $T-1$:

\begin{equation*}
    \begin{aligned}
        \phi\left(\{{u}_{k}\}_{k=t-T}^{t}\right)
        =
\left\{
\begin{bmatrix}
m_{p,k} \\
\sin(\theta_{p,k}^1) \\
\sin(2\theta_{p,k}^1) \\
\sin(\theta_{p,k}^2) \\
\sin(2\theta_{p,k}^2) \\
m_{r,k} \\
\sin(\theta_{r,k}^1) \\
\sin(2\theta_{r,k}^1) \\
\sin(\theta_{r,k}^2) \\
\sin(2\theta_{r,k}^2)
\end{bmatrix}
\right\}_{k=t-T}^{t-2}.\\
    \end{aligned}
\end{equation*}
When computing $\mathcal{L}_{\text{invar}}$, we transform two types of trajectories: 1) $\phi(\{\hat{u}_{k}\}_{k=t-T}^{t})$, the transformed values from the demonstration trajectories, and 2) $\phi(u_t, \{\hat{u}_{k}\}_{k=t-T}^{t-1})$, the transformed values from the given previous trajectories $\{\hat{u}_{k}\}_{k=t-T}^{t-1}$ and the predicted target $u_t$ at time $t$. By calculating the L2 loss between these two transformed values and using it as a training loss, the predicted trajectories $u_t$ are aligned with the demonstration trajectories in the motion-invariant space, given $\{\hat{u}_{k}\}_{k=t-T}^{t-1}$.

\begin{figure*}[!t]
	\centering
	\includegraphics[width=\linewidth]{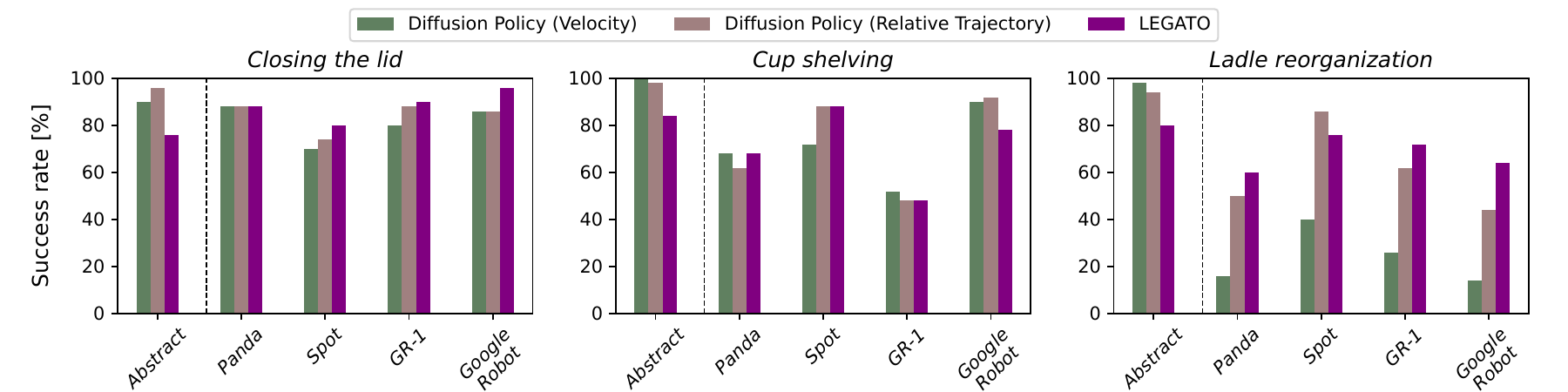}
	\caption{
    \textbf{Comparison with different types of Diffusion Policy baselines}    
        We report success rates from 50 trials of our LEGATO policies compared to Diffusion Policy baselines, using different action spaces: end-effector velocities~\cite{chi2024diffusionpolicy} and relative end-effector trajectories~\cite{chi2024umi}.
        Although Diffusion Policy (Relative Trajectory) outperformed Diffusion Policy (Velocity) in cross-embodiment settings, \ourmethod{} policies achieved the highest success rates across most cross-embodiment settings, except for deployment on the \textit{Goodle Robot} for the \textit{Cup shelving} task and the \textit{Spot} robot for the \textit{Ladle reorganization} task.
    }
    \vspace{2pt}
	\label{fig:diffusion}
\end{figure*}

\subsection{Supplementary Evaluation in Simulation}
We provide additional quantitative evaluations in simulation to further discuss cross-embodiment visuomotor policies. Specifically, we aim to address: 1) a comparison with existing cross-embodiment learning frameworks that utilize different action spaces, and 2) the applicability of the motion-invariant regularization to varied neural network architectures and its impact on their performance.

\paragraph{Comparison with the Diffusion Policy Using Relative-Trajectory Actions}
To compare our method with existing works aimed at cross-embodiment learning of ego-centric visuomotor policies, we adapted the Diffusion Policy with the action space of relative end-effector trajectories, as used in Universal Manipulation Interface \cite{chi2024umi}, to our setting, where the visuomotor policy outputs gripper trajectories based on previous actions and visual observations.
We refer to this baseline as Diffusion Policy (Relative Trajectory). This baseline serves as a reference for Universal Manipulation Interface \cite{chi2024umi}, but without the latency compensation process.
As described in Section~\ref{sec:intro}, generalizing latency compensation across various robot embodiments is challenging because it requires fine-tuning for each target robot system. Therefore, we exclude the latency compensation process in our evaluation.

We used the same setup as the quantitative evaluation in Section~\ref{sec:sim-eval}, utilizing the same dataset of 150 task demonstrations with the \textit{Abstract} embodiment.
In Figure~\ref{fig:diffusion}, we report the success rates for deploying Diffusion Policy (Relative Trajectory) with \ourmethod{} and the Diffusion Policy baseline used in Section~\ref{sec:sim-eval}. 
For clarification, the Diffusion Policy baseline from Section~\ref{sec:sim-eval} is referred to as Diffusion Policy (Velocity).
In our work, we considered only the action space of the handheld gripper's differential poses to minimize the impact of visual odometry errors during demonstration collection and to eliminate reliance on specified frames other than the handheld gripper's pose.
This ensures suitability across various robot platforms, including floating-base robot systems.
Unlike our setting, training with the action space of relative trajectories requires additional geometric information, such as the initial gripper pose for a sequence of receding-horizon actions, to generate the training dataset.
This information is not incorporated into other baselines or the \ourmethod{} framework.
Nevertheless, \ourmethod{} achieved higher performance than Diffusion Policy (Relative Trajectory) in cross-embodiment settings for the \textit{Closing the lid} and \textit{Ladle reorganization} tasks, though not in deployment within the same domain. 
As noted by Chi et al. \cite{chi2024umi}, Diffusion Policy (Relative Trajectory) outperformed Diffusion Policy (Velocity) in their evaluation, and this was also observed in our setting.

\begin{figure*}[!t]
	\centering
	\includegraphics[width=\linewidth]{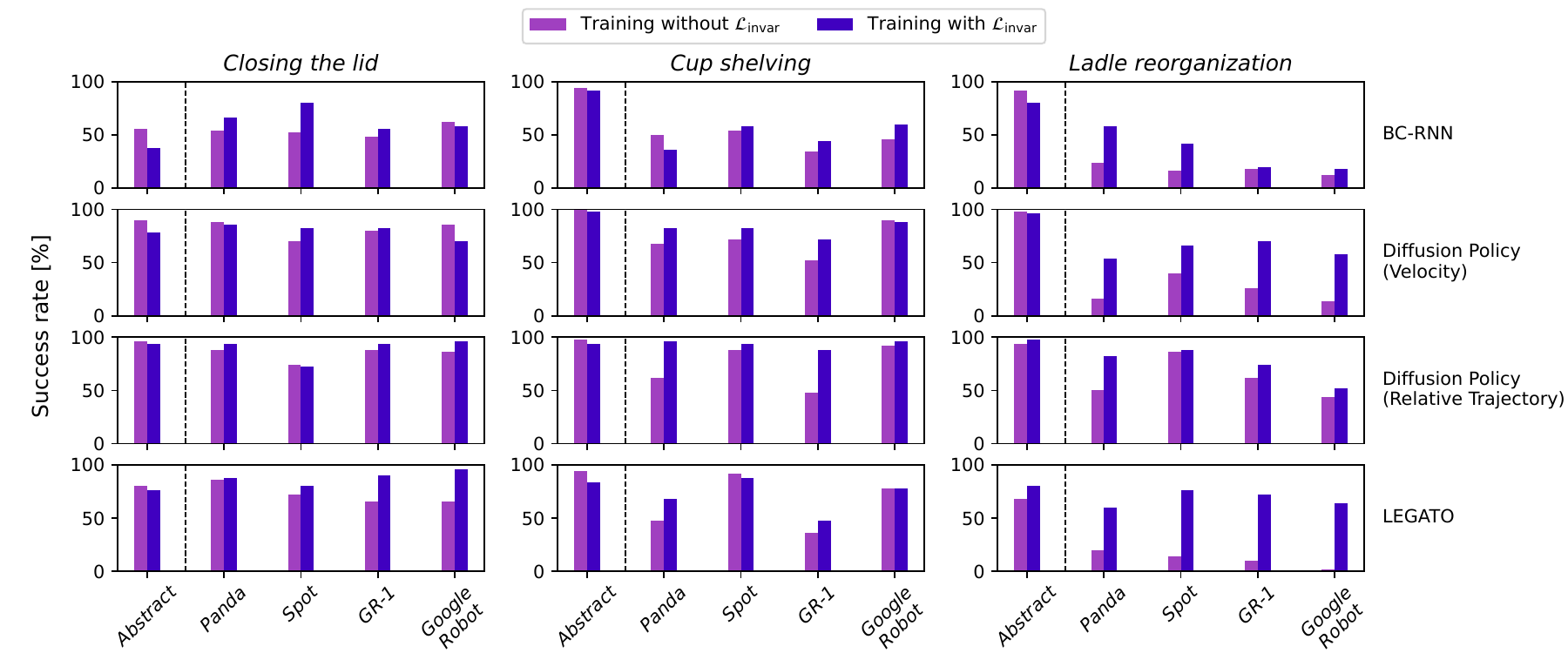}
	\caption{
    \textbf{Impact of training with the motion-invariant regularization}
        We evaluate the success rates of policies trained with and without the motion-invariant regularization, $\mathcal{L}_{\text{invar}}$, over 50 trials across varied architectures. Each policy is trained on demonstrations from the \textit{Abstract} embodiment.
        While the success rates on the \textit{Abstract} embodiment (used for demonstrations) decrease with the motion-invariant regularization, the success rates improve in cross-embodiment settings, where policies are deployed to other robots.
    }
	\label{fig:invariant}
\end{figure*}

\paragraph{Impact of Training with the Motion-Invariant Regularization}
We provide an additional ablation study on the complementary use of the motion-invariant regularization loss to enhance cross-embodiment transferability in policies, as shown in Figure~\ref{fig:invariant}. Specifically, we applied the motion-invariant regularization loss $\mathcal{L}_{\text{invar}}$, as described in Section~\ref{sec:policy} and Appendix~\ref{sec:invar}, to BC-RNN \cite{mandlekar2021matters}, Diffusion Policy (Velocity) adapted from the Velocity Diffusion Policy introduced by Chi et al. \cite{chi2024diffusionpolicy}, and Diffusion Policy (Relative Trajectory) adapted from the Diffusion Policy using relative end-effector trajectories as the action space in Universal Manipulation Interface \cite{chi2024umi}.
Similar to the quantitative evaluation in Section~\ref{sec:sim-eval}, all baselines are trained on the same dataset of 150 task demonstrations using the \textit{Abstract} embodiment. Additionally, the same IK optimizer with consistent parameters is used within a single robot.

As outlined in Equation~\ref{eq:loss} of Section~\ref{sec:policy}, the motion-invariant regularization loss are added to the original loss functions for each baseline. In BC-RNN, the motion-invariant regularization loss is added to the negative log-likelihood between the predicted and demonstration actions. For the Diffusion Policy baselines, the motion-invariant regularization loss is integrated with the L2-based DDPM loss during the denoising process. The motion-invariant regularization loss is adapted for the sequence of receding-horizon actions used in the Diffusion Policy baselines and is defined as:
\begin{equation*}
\mathcal{L}_{\text{invar}} = \sum \lVert
\phi(\{u_{k}\}_{k=t}^{t+P}, \{\hat{u}_{k}\}_{k=t-T}^{t-1}) - \phi(\{\hat{u}_{k}\}_{k=t-T}^{t+P})
\rVert ^2,
\end{equation*}
where $P$ is the prediction horizon, and the sequence of actions of length $P+1$ is transformed into the motion-invariant space.

Our findings indicate that incorporating the motion-invariant regularization during training generally reduces success rates when deploying policies on the \textit{Abstract} embodiment but enhances performance in cross-embodiment settings with different robot embodiments, regardless of the neural network architectures.
This highlights the applicability of leveraging motion invariance across various neural network architectures and its effectiveness for cross-embodiment learning.

\subsection{Implementation Details}
The visuomotor policy $\pi_{H}$ predicts target poses for the handheld gripper at 10 Hz. The IK optimizer $\pi_{L}$ realizes these target poses by retargeting them into whole-body motions, updating target joint positions and body orientation at 100 Hz. In simulation, we applied low-level PD control for each joint and body at 500 Hz. For the \textit{Spot} robot, we additionally computed joint positions for the legs by solving IK analytically based on the target body pose. For the \textit{Google Robot}, body motion was controlled similarly to other arm joints with PD control, though using high gains. In real robot setups, we controlled the robots through APIs provided by the manufacturers. For quantitative evaluation on the \textit{Panda} robot, we used \texttt{JOINT\_IMPEDANCE} mode via Deoxys~\cite{zhu2022viola} for joint position control. In the demonstration on the \textit{Spot} robot, we directly streamed one-point trajectories for arm joint positions and body poses through Boston Dynamics' Spot SDK.

\subsection{Demonstrations in Simulation}
Task demonstrations in simulation use the same tracking camera setup as in real-robot evaluations—a Realsense T265~\cite{realsense2022} camera.
To replicate real-world human demonstration behaviors, visual odometry data from the tracking camera is mapped to simulated handheld gripper motions in the \textit{Abstract} embodiment or to IK commands for teleoperated simulation robots. 
The button interface for triggering grasp actions and recording data is kept consistent with the real-world setup. 
However, unlike real-world demonstrations, simulation does not require physical interaction with the handheld gripper. 
Therefore, shared gripper components were removed, and a simplified handle was used to reduce the workload on the human demonstrators.
\newpage
\null

%% file: references.bib
@inproceedings{kemp2022stretch,
  title={The design of stretch: A compact, lightweight mobile manipulator for indoor human environments},
  author={Kemp, Charles C and Edsinger, Aaron and Clever, Henry M and Matulevich, Blaine},
  booktitle={IEEE International Conference on Robotics and Automation},
  year={2022},
}

@MISC{bd-spot,
title =    {{Boston Dynamics} {Spot}},
howpublished = {\url{https://bostondynamics.com/products/spot}},
}

@MISC{unitree-b1,
title =    {Unitree {B1}},
howpublished = {\url{https://shop.unitree.com/products/unitree-b1}},
}

@MISC{fourier-gr1,
title =    {{GR1} - {Fourier Intelligence}},
howpublished = {\url{https://fourierintelligence.com/gr1}},
}

@MISC{agility-digit,
title =    {{Agility Robotics} {Digit}},
howpublished = {\url{https://agilityrobotics.com/robots}},
}

@inproceedings{xie2020deep,
  title={Deep imitation learning for bimanual robotic manipulation},
  author={Xie, Fan and Chowdhury, Alexander and De Paolis Kaluza, M and Zhao, Linfeng and Wong, Lawson and Yu, Rose},
  booktitle={Advances in Neural Information Processing Systems},
  year={2020}
}

@inproceedings{seo2023trill,
   title={Deep Imitation Learning for Humanoid Loco-manipulation through Human Teleoperation},
   author={Seo, Mingyo and Han, Steve and Sim, Kyutae and 
           Bang, Seung Hyeon and Gonzalez, Carlos and 
           Sentis, Luis and Zhu, Yuke},
   booktitle={IEEE-RAS International Conference on Humanoid Robots},
   year={2023}
}

@inproceedings{zhao2023aloha,
   title={Learning fine-grained bimanual manipulation with low-cost hardware},
   author={Zhao, Tony Z and Kumar, Vikash and Levine, Sergey and Finn, Chelsea},
   booktitle={Robotics: Science and Systems},
   year={2023}
}

@inproceedings{mandlekar2023mimicgen,
  title={Mimicgen: A data generation system for scalable robot learning using human demonstrations},
  author={Mandlekar, Ajay and Nasiriany, Soroush and Wen, Bowen and Akinola, Iretiayo and Narang, Yashraj and Fan, Linxi and Zhu, Yuke and Fox, Dieter},
  booktitle={Conference on Robot Learning},
  year={2023}
}

@inproceedings{sridhar2023memory,
  title={Memory-consistent neural networks for imitation learning},
  author={Sridhar, Kaustubh and Dutta, Souradeep and Jayaraman, Dinesh and Weimer, James and Lee, Insup},
  booktitle={International Conference on Learning Representations},
  year={2024}
}

@article{song2020grasping,
  title={Grasping in the wild: Learning 6dof closed-loop grasping from low-cost demonstrations},
  author={Song, Shuran and Zeng, Andy and Lee, Johnny and Funkhouser, Thomas},
  journal={IEEE Robotics and Automation Letters},
  year={2020},
}

@inproceedings{young2021visual,
  title={Visual imitation made easy},
  author={Young, Sarah and Gandhi, Dhiraj and Tulsiani, Shubham and Gupta, Abhinav and Abbeel, Pieter and Pinto, Lerrel},
  booktitle={Conference on Robot Learning},
  year={2021},
}

@inproceedings{seo2022prelude,
        title={Learning to Walk by Steering: Perceptive Quadrupedal Locomotion
          in Dynamic Environments},
        author={Seo, Mingyo and Gupta, Ryan and Zhu, Yifeng and Skoutnev, Alexy
          and Sentis, Luis and Zhu, Yuke},
        booktitle={IEEE International Conference on Robotics and Automation},
        year={2023}
      }

@inproceedings{chi2024umi,
  title={Universal Manipulation Interface: In-The-Wild Robot Teaching Without In-The-Wild Robots},
  author={Chi, Cheng and Xu, Zhenjia and Pan, Chuer and Cousineau, Eric and Burchfiel, Benjamin and Feng, Siyuan and Tedrake, Russ and Song, Shuran},
   booktitle={Robotics: Science and Systems},
  year={2024}
}

@inproceedings{smith2019avid,
   title={Avid: Learning multi-stage tasks via pixel-level translation of human videos},
   author={Smith, Laura and Dhawan, Nikita and Zhang, Marvin and Abbeel, Pieter and Levine, Sergey},
   booktitle={Robotics: Science and Systems},
   year={2020}
}

@inproceedings{zakka2022xirl,
  title={Xirl: Cross-embodiment inverse reinforcement learning},
  author={Zakka, Kevin and Zeng, Andy and Florence, Pete and Tompson, Jonathan and Bohg, Jeannette and Dwibedi, Debidatta},
  booktitle={Conference on Robot Learning},
  year={2022},
}

@inproceedings{shaw2023videodex,
  title={Videodex: Learning dexterity from internet videos},
  author={Shaw, Kenneth and Bahl, Shikhar and Pathak, Deepak},
  booktitle={Conference on Robot Learning},
  year={2023},
}

@inproceedings{xu2023xskill,
  title={Xskill: Cross embodiment skill discovery},
  author={Xu, Mengda and Xu, Zhenjia and Chi, Cheng and Veloso, Manuela and Song, Shuran},
  booktitle={Conference on Robot Learning},
  year={2023},
}

@inproceedings{okami2024,
    title={OKAMI: Teaching Humanoid Robots Manipulation Skills through Single Video Imitation},
    author={Jinhan Li and Yifeng Zhu and Yuqi Xie and Zhenyu Jiang and Mingyo Seo and Georgios Pavlakos and Yuke Zhu},
    booktitle={Conference on Robot Learning},
    year={2024}
}

@inproceedings{
    fang2023airexo,
    title = {Low-Cost Exoskeletons for Learning Whole-Arm Manipulation in the Wild},
    author = {Fang, Hongjie and Fang, Hao-Shu and Wang, Yiming and Ren, Jieji and Chen, Jingjing and Zhang, Ruo and Wang, Weiming and Lu, Cewu},
    booktitle={IEEE International Conference on Robotics and Automation},
    year = {2024}
}

@article{kim2023training,
  title={Training robots without robots: deep imitation learning for master-to-robot policy transfer},
  author={Kim, Heecheol and Ohmura, Yoshiyuki and Nagakubo, Akihiko and Kuniyoshi, Yasuo},
  journal={IEEE Robotics and Automation Letters},
  year={2023},
}

@inproceedings{ha2024umilegs,
      title={{UMI} on Legs: Making Manipulation Policies Mobile with Manipulation-Centric Whole-body Controllers}, 
      author={Huy Ha and Yihuai Gao and Zipeng Fu and Jie Tan and Shuran Song},
    booktitle={Conference on Robot Learning},
    year={2024}
}

@article{tachi2003telexistence,
  title={Telexistence cockpit for humanoid robot control},
  author={Tachi, Susumu and Komoriya, Kiyoshi and Sawada, Kazuya and Nishiyama, Takashi and Itoko, Toshiyuki and Kobayashi, Masami and Inoue, Kozo},
  journal={Advanced Robotics},
  year={2003},
}

@article{ayusawa2017motion,
  title={Motion retargeting for humanoid robots based on simultaneous morphing parameter identification and motion optimization},
  author={Ayusawa, Ko and Yoshida, Eiichi},
  journal={IEEE Transactions on Robotics},
  year={2017},
}

@inproceedings{penco2018robust,
  title={Robust real-time whole-body motion retargeting from human to humanoid},
  author={Penco, Luigi and Cl{\'e}ment, Brice and Modugno, Valerio and Hoffman, E Mingo and Nava, Gabriele and Pucci, Daniele and Tsagarakis, Nikos G and Mouret, J-B and Ivaldi, Serena},
  booktitle={IEEE-RAS International Conference on Humanoid Robots},
  year={2018},
}

@inproceedings{darvish2019whole,
  title={Whole-body geometric retargeting for humanoid robots},
  author={Darvish, Kourosh and Tirupachuri, Yeshasvi and Romualdi, Giulio and Rapetti, Lorenzo and Ferigo, Diego and Chavez, Francisco Javier Andrade and Pucci, Daniele},
  booktitle={IEEE-RAS International Conference on Humanoid Robots},
  year={2019},
}

@inproceedings{
	RoboImitationPeng20,
	author = {Peng, Xue Bin and Coumans, Erwin and Zhang, Tingnan and Lee, Tsang-Wei Edward and Tan, Jie and Levine, Sergey},
	booktitle={Robotics: Science and Systems},
	year = {2020},
	title = {Learning Agile Robotic Locomotion Skills by Imitating Animals},
}

@inproceedings{
	pari2022vinn,
	author = {Pari, Jyothish and Shafiullah, Nur Muhammad and Arunachalam, Sridhar Pandian and Pinto, Lerrel},
	booktitle={Robotics: Science and Systems},
	year = {2022},
	title = {The Surprising Effectiveness of Representation Learning for Visual Imitation},
}

@inproceedings{he2024learning,
  title     = {Learning Human-to-Humanoid Real-Time Whole-Body Teleoperation},
  author    = {He, Tairan and Luo, Zhengyi and Xiao, Wenli and Zhang, Chong and Kitani, Kris and Liu, Changliu and Shi, Guanya},
  booktitle={IEEE/RSJ International Conference on Intelligent Robots and Systems},
  year={2024}
}

@inproceedings{fu2024humanplus,
  author    = {Fu, Zipeng and Zhao, Qingqing and Wu, Qi and Wetzstein, Gordon and Finn, Chelsea},
  title     = {HumanPlus: Humanoid Shadowing and Imitation from Humans},
  booktitle = {Conference on Robot Learning},
  year      = {2024},
}

@inproceedings{kim2022human,
  title={Human motion control of quadrupedal robots using deep reinforcement learning},
  author={Kim, Sunwoo and Sorokin, Maks and Lee, Jehee and Ha, Sehoon},
  booktitle={Robotics: Science and Systems},
  year={2022}
}

@inproceedings{li2023ace,
  title={Ace: Adversarial correspondence embedding for cross morphology motion retargeting from human to nonhuman characters},
  author={Li, Tianyu and Won, Jungdam and Clegg, Alexander and Kim, Jeonghwan and Rai, Akshara and Ha, Sehoon},
  booktitle={SIGGRAPH Asia},
  year={2023}
}

@article{saveriano2023dmp,
  title={Dynamic movement primitives in robotics: A tutorial survey},
  author={Saveriano, Matteo and Abu-Dakka, Fares J and Kramberger, Alja{\v{z}} and Peternel, Luka},
  journal={The International Journal of Robotics Research},
  year={2023},
}

@inproceedings{paraschos2013promp,
  title={Probabilistic movement primitives},
  author={Paraschos, Alexandros and Daniel, Christian and Peters, Jan R and Neumann, Gerhard},
  booktitle={Advances in Neural Information Processing Systems},
  year={2013}
}

@inproceedings{zhou2019vmp,
  title={Learning via-point movement primitives with inter-and extrapolation capabilities},
  author={Zhou, You and Gao, Jianfeng and Asfour, Tamim},
  booktitle={IEEE/RSJ International Conference on Intelligent Robots and Systems},
  year={2019},
}

@inproceedings{noseworthy2020tc-vae,
  title={Task-conditioned variational autoencoders for learning movement primitives},
  author={Noseworthy, Michael and Paul, Rohan and Roy, Subhro and Park, Daehyung and Roy, Nicholas},
  booktitle={Conference on robot learning},
  year={2020},
}

@inproceedings{urain2020imitationflow,
  title={Imitationflow: Learning deep stable stochastic dynamic systems by normalizing flows},
  author={Urain, Julen and Ginesi, Michele and Tateo, Davide and Peters, Jan},
  booktitle={IEEE/RSJ International Conference on Intelligent Robots and Systems},
  year={2020},
}

@inproceedings{lee2023emmp,
  title={Equivariant motion manifold primitives},
  author={Lee, Byeongho and Lee, Yonghyeon and Kim, Seungyeon and Son, MinJun and Park, Frank C},
  booktitle={Conference on Robot Learning},
  year={2023}
}

@inproceedings{luo2023pulse,
  title={Universal Humanoid Motion Representations for Physics-Based Control},
  author={Luo, Zhengyi and Cao, Jinkun and Merel, Josh and Winkler, Alexander and Huang, Jing and Kitani, Kris and Xu, Weipeng},
  booktitle={International Conference on Learning Representations},
  year={2024}
}

@inproceedings{yan2023imitationnet,
  title={{ImitationNet}: Unsupervised Human-to-Robot Motion Retargeting via Shared Latent Space},
  author={Yan, Yashuai and Mascaro, Esteve Valls and Lee, Dongheui},
  booktitle={IEEE-RAS International Conference on Humanoid Robots},
  year={2023},
}

@article{lee2018dhb,
  title={Bidirectional invariant representation of rigid body motions and its application to gesture recognition and reproduction},
  author={Lee, Dongheui and Soloperto, Raffaele and Saveriano, Matteo},
  journal={Autonomous Robots},
  year={2018},
}

@article{noguchi2021tool,
  title={Tool as embodiment for recursive manipulation},
  author={Noguchi, Yuki and Matsushima, Tatsuya and Matsuo, Yutaka and Gu, Shixiang Shane},
  journal={arXiv preprint arXiv:2112.00359},
  year={2021}
}

@inproceedings{wang2020critic,
  title={Critic regularized regression},
  author={Wang, Ziyu and Novikov, Alexander and Zolna, Konrad and Merel, Josh S and Springenberg, Jost Tobias and Reed, Scott E and Shahriari, Bobak and Siegel, Noah and Gulcehre, Caglar and Heess, Nicolas and others},
  booktitle={Advances in Neural Information Processing Systems},
  year={2020}
}

@article{fiore2023general,
  title={A general framework for hierarchical redundancy resolution under arbitrary constraints},
  author={Fiore, Mario Daniele and Meli, Gaetano and Ziese, Anton and Siciliano, Bruno and Natale, Ciro},
  journal={IEEE Transactions on Robotics},
  year={2023},
}

@article{hochreiter1997long,
  title={Long short-term memory},
  author={Hochreiter, Sepp and Schmidhuber, J{\"u}rgen},
  journal={Neural Computation},
  year={1997},
  publisher={MIT Press}
}

@inproceedings{he2016deep,
  title={Deep residual learning for image recognition},
  author={He, Kaiming and Zhang, Xiangyu and Ren, Shaoqing and Sun, Jian},
  booktitle={IEEE Conference on Computer Vision and Pattern Recognition},
  year={2016}
}

@inproceedings{todorov2012mujoco,
  title={MuJoCo: A physics engine for model-based control},
  author={Todorov, Emanuel and Erez, Tom and Tassa, Yuval},
  booktitle={IEEE/RSJ International Conference on Intelligent Robots and Systems},
  year={2012},
}

@MISC{realsense2022,
title =    {Intel {RealSense SDK}},
howpublished = {\url{https://github.com/IntelRealSense/librealsense}},
}

@MISC{seersense2024,
title =    {{SeerSense™} {XR50 Module}},
howpublished = {\url{https://www.xvisiotech.com/product/seersense-xr50/}},
}

@MISC{franka-panda,
title =    {{Franka Robotics}},
howpublished = {\url{https://franka.de/}},
}

@inproceedings{mandlekar2021matters,
  title={What Matters in Learning from Offline Human Demonstrations for Robot Manipulation},
  author={Mandlekar, Ajay and Xu, Danfei and Wong, Josiah and Nasiriany, Soroush and Wang, Chen and Kulkarni, Rohun and Fei-Fei, Li and Savarese, Silvio and Zhu, Yuke and Mart{\'\i}n-Mart{\'\i}n, Roberto},
  booktitle={Conference on Robot Learning},
  year={2022},
}

@article{chi2024diffusionpolicy,
	author = {Cheng Chi and Zhenjia Xu and Siyuan Feng and Eric Cousineau and Yilun Du and Benjamin Burchfiel and Russ Tedrake and Shuran Song},
	title ={Diffusion Policy: Visuomotor Policy Learning via Action Diffusion},
	journal = {The International Journal of Robotics Research},
	year = {2024},
}

@inproceedings{herzog2023deep,
  title={Deep RL at scale: Sorting waste in office buildings with a fleet of mobile manipulators},
  author={Herzog, Alexander and Rao, Kanishka and Hausman, Karol and Lu, Yao and Wohlhart, Paul and Yan, Mengyuan and Lin, Jessica and Arenas, Montserrat Gonzalez and Xiao, Ted and Kappler, Daniel and others},
  booktitle={Robotics: Science and Systems},
  year={2023}
}

@inproceedings{wang2023mimicplay,
    title={Mimicplay: Long-horizon imitation learning by watching human play},
    author={Wang, Chen and Fan, Linxi and Sun, Jiankai and Zhang, Ruohan and Fei-Fei, Li and Xu, Danfei and Zhu, Yuke and
    Anandkumar, Anima},
    booktitle={Conference on Robot Learning},
    year={2023}
}

@inproceedings{zhu2022viola,
  title={Viola: Imitation learning for vision-based manipulation with object proposal priors},
  author={Zhu, Yifeng and Joshi, Abhishek and Stone, Peter and Zhu, Yuke},
  booktitle={Conference on Robot Learning},
  year={2022}
}
